\documentclass[runningheads,a4paper]{llncs}

\usepackage{amsmath,amsfonts,amssymb}
\usepackage{graphicx}
\usepackage{algorithm}
\usepackage{algorithmicx}
\usepackage{algpseudocode}
\usepackage{multirow}
\usepackage{mathtools}
\usepackage{chngpage}

\usepackage{xcolor}
\usepackage[normalem]{ulem}

\usepackage{hyperref}

\definecolor{dark-red}{rgb}{0.4,0.15,0.15}
\definecolor{dark-blue}{rgb}{0.15,0.15,0.4}
\definecolor{medium-blue}{rgb}{0.0,0.0,0.4}
\hypersetup{
	colorlinks, linkcolor={black},
	citecolor={dark-blue}, urlcolor={medium-blue}
}

\newcommand{\comment}[1]{}
\newcommand{\argmin}{\operatornamewithlimits{argmin}}

\title{Discriminative Parameter Estimation\\
 for Random Walks Segmentation: \\
 Technical Report}
 
 \author{
    Pierre-Yves~Baudin\inst{1-6} \and 
    Danny~Goodman\inst{1-3} \and 
    Puneet~Kumar\inst{1-3} \and 
    Noura~Azzabou\inst{4-6} \and
    Pierre~G.~Carlier\inst{4-6} \and 
    Nikos~Paragios\inst{1-3} \and
    M.~Pawan~Kumar\inst{1-3}
    }

\institute{
    Center for Visual Computing, \'{E}cole Centrale Paris, FR \and 
    Universit\'{e} Paris-Est, LIGM (UMR CNRS), \'{E}cole des Ponts ParisTech, FR \and 
    \'{E}quipe Galen, INRIA Saclay, FR \and
    Institute of Myology, Paris, FR \and 
    CEA, I$^{2}$ BM, MIRCen, IdM NMR Laboratory, Paris, FR \and 
    UPMC University Paris 06, Paris, FR
    }

\authorrunning{Pierre-Yves~Baudin et al.}
\titlerunning{Parameter Estimation for RW Segmentation: \\
 Technical Report}

\begin{document}
\maketitle

\section{Literature Survey on the State of the Art Algorithms in Muscle Segmentation}

In the following, we present the principal methods addressing the segmentation of skeletal muscles. 

\paragraph{Muscle Segmentation using Simplex meshes with medial representations
\label{par:soa_gilles1}}

A skeletal muscle segmentation method was presented in~\cite{Gilles2010}
based on simplex meshes~\cite{delingette_these}. Considering a 3D
surface model, simplex meshes are discrete meshes where each vertex
has exactly 3 neighbors. Having a constant connectivity allows to
simply parametrize the location of one vertex with respect to its
neighbors, and thus parametrize deformation of the shape -- translation,
rotation, scaling -- in a local manner. Indeed, the location of a
pixel, denoted as ${\bf x}$ can be expressed as a linear combination
of the locations of the three neighbors plus a local elevation term
parallel to the local normal\comment{ (see figure~\ref{fig:soa_simplex})}:
${\bf x}=\varepsilon_{1}{\bf x}_{1}+\varepsilon_{2}{\bf x}_{2}+\left(1-\varepsilon_{1}-\varepsilon_{2}\right){\bf x}_{3}+h{\bf n}.$
As a result, many local measurements -- including curvature and cell
surface -- can be computed efficiently and global energy terms enforcing
local constraints come up naturally. 

Here, the authors impose local smoothing via curvature averaging,
which does not tend to reduce the surface like 1-order operators typically
do. Prior knowledge is imposed by constraining the local scale changes
on the elevation parameter with respect to a reference shape. Denoting
the surface of the triangle formed by the three neighbors of a pixel
as $S$, given the reference shape parameters $\left(\tilde{\varepsilon}_{1},\tilde{\varepsilon}_{2},\tilde{h},\tilde{S}\right)$,
the new location of the considered pixel is expressed as:
\begin{equation}
{\bf x}=\tilde{\varepsilon}_{1}{\bf x}_{1}+\tilde{\varepsilon}_{2}{\bf x}_{2}+\left(1-\tilde{\varepsilon}_{1}-\tilde{\varepsilon}_{2}\right){\bf x}_{3}+\tilde{h}\left(S/\tilde{S}\right)^{1/\beta}{\bf n},
\end{equation}
where $\beta\in[2,+\infty[$ is a parameter which sets the amount
of allowed local deformation: with $\beta=2$ this definition is similitude
invariant; with $\beta=+\infty$ this definition is invariant through
rigid transformations only. The model is attached to the target image
through either gradient norm maximization in the direction of the
gradient at the location of the vertices, or maximization of similarities
between the reference and the target images at the vertices location.

A medial representation -- similar to the M-reps~\cite{pizer2003deformable}
-- is combined with the simplex parametrization to exploit the specific
tubular shapes of the skeletal muscles. Medial vertices are added
to the model, constrained to remain on the medial axis of the tubular
objects. This is achieved by connecting the new vertices to the surface
vertices through spring-like forces. This constrains the global structure
to resemble its initial reference shape, thus acting as a global shape
prior\comment{(see figure~\ref{fig:soa_simplex})}. This medial axis representation
also allows efficient collision handling. The model is fit to the
image through an iterative process of successive local evolutions.
Such model appear to always yield a valid solution, sometimes at the
price of an excessive regularization or lack of adaptability to the
specifics of the target image.
paragraph{Muscle Segmentation using Deformable Models and Shape Matching\label{par:soa_gilles2}}

\paragraph{Muscle Segmentation using Deformable Models and Shape Matching
\label{par:soa_gilles2}}
A shape prior for muscle segmentation in 3D images
 was presented in~\cite{Gilles2008}, deriving from
a computer animation technique, called\emph{ shape matching}, used
to efficiently approximate large soft-tissue elastic deformations. This method was applied to muscle segmentation
with some success. In this approach, discrete meshes are used to parametrize
the moving surface. Let $\mathbf{x}^{0}$ be the vector containing
the initial position of the control points of the parametric surface.
Clustering is performed on $\mathbf{x}^{0}$ such that each cluster
$\zeta_{i}$ contains at least a certain number of vertices (set by
the user). During segmentation, the evolution of the active surface
is performed according to the following iterative procedure:
\begin{enumerate}
\item Shift vertices according to the external force: $\tilde{\mathbf{x}}^{t}=\mathbf{f}_{\mathrm{ext}}+\mathrm{\mathbf{x}}^{t}$.
The external ``force'' $\mathbf{f}_{\mathrm{ext}}$ is computed
as the maximal gradient search in the gradient direction.
\item Regularize vertex positions:

\begin{enumerate}
\item Compute rigid registration for each clusters:
\begin{align}
    \mathbf{T}_{i} = 
    \arg\min\sum_{ j\in\zeta_{i}}\left\Vert \mathbf{T}_{i}\mathbf{x}_{j}^{0} - 
        \tilde{\mathbf{x}}_{j}^{t}\right\Vert ^{2},
\end{align}
\item Average target position for each vertex:
\begin{align}
    \mathbf{x}_{i}^{t+1} = 
        \frac{1}{\left|\zeta_{i}\right|}\sum_{j\in\zeta_{i}}\mathbf{T}_{j}\mathbf{x}_{j}^{0}.
\end{align}
\end{enumerate}
\end{enumerate}
Single reference prior models are convenient in that they require
only one annotated example of the objects of interest. However, when
segmenting a class of objects whose shape varies a lot, such approach
becomes too constraining and does not allow the model to adopt valid
shapes which are too different from the single reference. 

\paragraph{Muscle segmentation using a hierarchical statistical shape model
\label{par:soa_essafi}}
A hierarchical prior model using Diffusion Wavelets was proposed to
segment organs in~\cite{Essafi2009} -- including one calf muscle --
in MRI. This model builds on the formulation of the ASM~\cite{Cootes1995}\comment{(cf. equation~\ref{eq:ASM})},
using a different basis for the subspace of valid solutions. One of
the main drawbacks of ASMs, is that they often require a large number
of training data in order to obtain relevant decomposition modes.
Indeed, some non-correlated shape features -- such as global and local
shape configurations -- are often modeled by the same deformation
modes. Thus, desired shape behaviors are often mixed with unwanted
shape behaviors when optimizing the shape parameters for segmenting
a new image. The hierarchical approach allows to uncorrelate small
and large scale shape behaviors. Moreover, the presented method also
uncorrelates long-range shape behaviors, thus ensuring that deformation
mode are spatially localized.

We give a brief summary of this method. First, a graph $\mathcal{G}\left(\mathcal{V},\mathcal{E}\right)$
is built on the set of landmarks: $\mathcal{V}$ is the set of nodes
and each landmark corresponds to a node in $\mathcal{V}$; $\mathcal{E}$
is the set of edges, whose weights are determined through a statistical
analysis of the mutual relations between the landmarks in the training
set $\left\{ \mathbf{x}_{k}\right\} _{k=1\ldots K}$ (cf. Shape Maps
\cite{Langs2008}). As a result, landmarks with independent behaviors
will be connected by edges with a small weight, whereas nodes with
strongly related behaviors -- such as neighboring points -- will be
connected by large weight edges.

Second, a Diffusion Wavelet decomposition of $\mathcal{G}$ is performed.
This process involves computing the diffusion operator $\mathbf{T}$
of graph $\mathcal{G}$, which is the symmetric normalized Laplace-Beltrami
operator, and computing and compressing the dyadic powers of $\mathbf{T}$.
The output of this decomposition is a hierarchical orthogonal basis
$\left\{ \Gamma_{i}\right\} $ for the graph, whose vectors correspond
to different graph scales; considering the vector of landmark positions
when decomposed on the new basis: 
\begin{equation}
\mathbf{x}=\bar{\mathbf{x}}+\boldsymbol{\Gamma}\mathbf{p},
\end{equation}
global deformations -- i.e. global relations between all the nodes
-- are controlled by some of the coefficients in $\mathbf{p}$, while
local interactions -- i.e. local interactions between close-by nodes
-- are controlled by some other coefficients in $\mathbf{p}$. Projecting
all the training examples onto this new basis, a PCA is performed
\emph{at each scale} of the decomposition. Finally, during the segmentation
process, the landmarks are positioned on the target image in an iterative
manner: 1) the position of the landmarks is updated according to a
local appearance model; 2) they are projected into the hierarchical
subspace defined previously.

\paragraph{Muscle segmentation using a continuous region model
\label{par:soa_andrews}}
A region-based segmentation method, proposed in~\cite{Cremers2008}, was extended to multi-label segmentation in~\cite{Andrews2011a},
and applied to skeletal muscle segmentation \cite{Andrews2011}.
Before performing the PCA on the training samples, an Isometric Log-Ratio
(ILR) transform is applied to the assignment vectors. The reason for
using this transform is that multi-label segmentation requires to
have probabilities at all times, which the previous method does not
achieve. Here, the PCA is performed in the ILR space and its output
is projected back into the initial probability space. Denoting $\eta_{\gamma}=\mu+\Gamma\gamma$
a segmentation in the subspace of valid solution spanned by the PCA
in the ILR space, the following functional is proposed:
\begin{equation}
E\left(\eta_{\gamma}\right)=d\left(\eta_{BG,}\eta_{\gamma}\right)^{2}+\int\left(1-h\left(x\right)\right)\left|\nabla\eta_{\gamma}\right|^{2}+\gamma^{T}\Sigma^{-1}\gamma,
\end{equation}
where $d\left(\eta_{BG},\cdot\right)$ is an intensity prior functional
for separating muscle voxels from background voxels, and $h\left(x\right)$
is and edge-map of the target image such that the energy is minimal
when the boundaries of the model match the edges in the image.

\section{The Random Walks Algorithm} 
    The Random Walks algorithm is graph-based: consider a graph 
    $\mathcal{G}=(\mathcal{V}, \mathcal{E})$, where $\mathcal{V}$ is a set of nodes -- corresponding to each voxel in the 3d image -- and $\mathcal{E}$ is a set of edges -- one per pair of adjacent voxels. Let us also denote a set of labels -- one per object to segment -- as $\mathcal{S}$.
    
    The aim of the Random Walks Algorithm~\cite{Grady2006} is to compute an assignment probability
    of all voxels to all labels. These probabilities depend on: 
    i) contrast between adjacent voxels, ii) manual -- thus deterministic -- assignments of some voxels, called \emph{seeds}, and iii) prior assignment probabilities. 
    
    The probabilities, contained in vector $\mathbf{y}$, can be obtained by minimizing the 
    following functional:
    
    \begin{align*}
      E_{\mathrm{RWprior}}(\mathbf{x}, \mathbf{y}) &=
          \mathbf{y}^{\top} L \mathbf{y} + 
          w \| \mathbf{y} - \mathbf{y}_0 \|^2_{\Omega({\mathbf{x}})} \\
          &=
          \mathbf{y}^{\top} \left[ 
              \sum_{\alpha} w_{\alpha} L_{\alpha}
                  \right] \mathbf{y}
          + 
          \sum_{\beta} w_{\beta} 
              \| \mathbf{y} - \mathbf{y}_{\beta} \|^2_{\Omega_{\beta}},
    \end{align*}
    which is a linear combination of Laplacians and prior terms. 
    
    The Laplacian matrices $L_{\alpha}$ contain the contrast terms. Its entries are of the form:
     \begin{equation}
        L_{i,j}=\begin{cases}
        \sum_{k}\omega_{kj} & \text{if }i=j,\\
        -\omega_{ij} & \text{if }(i,j)\in\mathcal{E},\\
        0 & \text{otherwise}.
        \end{cases}
      \end{equation}
    Here, $\omega_{ij}$ designates the weight of edge $(i,j)$. 
    It is usually computed as follows:
    
    \begin{equation}
        \omega_{ij}=\exp\left(-\beta\left(I_{i}-I_{j}\right)^{2}\right),
    \end{equation}
    where $I_i$ is the intensity of voxel $i$.
    
    In our experiments, we used three different Laplacians using this formnulation, with 
    three values of $\beta$: 50, 100 and 150 (with the image voxel values normalized 
    with their empirical standard deviation).
    
    We also implemented the lesser used alternate formulation:
    
    \begin{align}
        w_{ij}=\frac{1}{\beta\left|I_{i}-I_{j}\right|+\varepsilon} \,,
    \end{align}
    which we employed in one additional Laplacian term with $\beta=100$ and $\varepsilon=1$ 
    for comparison purposes, since the selected values do give good results on their own.    
    
    Since the objective function is quadratic in $\mathbf{y}$, its minimum can be computed by minimizing a linear system. The quadratic term, composed of a sum of Laplacians and diagonal matrices due to the prior term, is very sparse and has a specific structure due to the fact that only adjacent voxels are connected with an edge.
    
    Given the size of the problem (several millions of variables), this system has to be solved with iterative methods, such as Conjugate Gradient. The specific structure of the problem and the existence of parallelized algorithms (such as multigrid Conjugate Gradient) allow for an efficient optimization. For instance, our own implementation takes less than 20s for volumes of size $200\times200\times100$ on a regular desktop machine.
   
\section{Derivation of the Latent SVM Upper Bound}

Given a dataset ${\cal D}=\left\{ \left({\bf x}_{k},{\bf z}_{k}\right),\, k=1,\ldots,N\right\} $,
which consists of inputs ${\bf x}_{k}$ and their hard segmentation
${\bf z}_{k},$ we would like to estimate parameters ${\bf w}$ such
that the resulting inferred segmentations are accurate. Here, the
accuracy is measured using the loss function $\Delta\left(\cdot,\cdot\right)$.
Formally, let $\tilde{{\bf y}}_{k}\left({\bf w}\right)$ denote the
soft segmentation obtained by minimizing the energy functional $E\left(\cdot,{\bf x}_{k};\,{\bf w}\right)$
for the $k$-th training sample, that is, 
\begin{equation}
\tilde{{\bf y}}_{k}\left({\bf w}\right)=\arg\min_{{\bf y}}{\bf w}^{\top}\psi\left({\bf x}_{k},{\bf y}\right).\label{eq:learn_min_energy}
\end{equation}

We would like to learn the parameters ${\bf w}$ such that the empirical
risk is minimized over all samples in the dataset. In other words,
we would like to estimate the parameters ${\bf w}^{\star}$ such that
\begin{equation}
{\bf w}^{\star}=\arg\min_{{\bf w}}\frac{1}{N}\sum_{k}\Delta\left({\bf z}_{k},\tilde{{\bf y}}_{k}\left({\bf w}\right)\right).
\end{equation}
The above objective function is highly non-convex in ${\bf w}$, which
makes it prone to bad local minimum solutions. To alleviate this deficiency,
the latent SVM formulation upper bounds the risk for a sample $\left({\bf x},{\bf z}\right)$
as follows: 
\begin{eqnarray}
\Delta\left({\bf z}_{k},\tilde{{\bf y}}_{k}\left({\bf w}\right)\right) & = & \Delta\left({\bf z}_{k},\tilde{{\bf y}}_{k}\left({\bf w}\right)\right)+{\bf w}^{\top}\left[\psi\left({\bf x}_{k},\tilde{{\bf y}}_{k}\left({\bf w}\right)\right)-\psi\left({\bf x}_{k},\tilde{{\bf y}}_{k}\left({\bf w}\right)\right)\right],\\
 & \leq & \min_{\Delta\left({\bf z}_{k},\hat{{\bf y}}\right)=0}{\bf w}^{\top}\psi\left({\bf x}_{k},\hat{{\bf y}}\right)\\
 &  & -\left[{\bf w}^{\top}\psi\left({\bf x}_{k},\tilde{{\bf y}}_{k}\left({\bf w}\right)\right)-\Delta\left({\bf z}_{k},\tilde{{\bf y}}_{k}\left({\bf w}\right)\right)\right],\nonumber \\
 & \leq & \min_{\Delta\left({\bf z}_{k},\hat{{\bf y}}\right)=0}{\bf w}^{\top}\psi\left({\bf x}_{k},\hat{{\bf y}}\right)\\
 &  & -\min_{\overline{{\bf y}}}\left[{\bf w}^{\top}\psi\left({\bf x}_{k},\overline{{\bf y}}\right)-\Delta\left({\bf z}_{k},\overline{{\bf y}}\right)\right].\nonumber 
\end{eqnarray}
The first inequality follows from the fact that the prediction $\tilde{{\bf y}}_{k}\left({\bf w}\right)$
has the \emph{minimum possible energy} (see equation~\ref{eq:learn_min_energy}).
Thus, its energy has to be less than or equal to the energy of any
compatible segmentation $\hat{{\bf y}}$ with $\Delta\left({\bf z}_{k},{\bf \hat{{\bf y}}}\right)=0$.
The second inequality is true since it replaces the \emph{loss augmented}
energy of the prediction $\tilde{{\bf y}}_{k}\left({\bf w}\right)$
with the minimum loss augmented energy. 

This inequality leads to the following minimization problem:
\begin{gather}
\min_{{\bf w},\,\xi_{k}\geq0} 
    \lambda\left\Vert {\bf w}\right\Vert ^{2} + 
        \frac{1}{N}\sum_{k}\xi_{k},\label{eq:latent_SVM}\\
{\rm s.t.\ }
{\rm \min_{\Delta\left({\bf x}_{k},\hat{{\bf y}}\right)=0}}{\bf w}^{\top}\psi\left({\bf x}_{k},\hat{{\bf y}}\right)\leq{\bf w}^{\top}\psi\left({\bf x}_{k},\bar{{\bf y}}\right)-\Delta\left({\bf z}_{k},\bar{{\bf y}}\right)+\xi_{k},\,\forall\bar{{\bf y}},\,\forall k,\nonumber 
\end{gather} 
where $\lambda\left\Vert {\bf w}\right\Vert ^{2}$ is a regularization term, preventing overfitting the parameters to the training data.

\section{Dual-Decomposition Algorithm for the ACI}

Briefly, dual decomposition allows us to iteratively solve a convex optimization problem of the form	\vspace{-1em}
\begin{equation}
{\bf y}^* = \argmin_{{\bf y} \in {\cal F}} \sum_{m=1}^M g_m({\bf y}).
\end{equation}
At each iteration $t$ it solves a set of slaves problems	\vspace{-0.5em}
\begin{equation}
{\bf y}_m^* = \argmin_{{\bf y}_m \in {\cal F}} \left( g_m({\bf y}_m)  + \rho^t_m {\bf y}_m \right),
\end{equation}
where $\rho^t_m$ are the dual variables satisfying $\sum_m \rho^t_m = 0$. The dual variables are initialized as
$\rho^0_m = 0, \forall m$, and updated at iteration $t$ as follows:
\begin{equation}
\rho^{t+1}_m \leftarrow \rho^t_m + \eta^t ({\bf y}^*_m - \sum_n {\bf y}^*_n/M),
\end{equation}
where $\eta^t$ is the learning rate at iteration $t$.
Under fairly general conditions, this iterative strategy
converges to the globally optimal solution of the original problem, that is,
${\bf y}^* = {\bf y}^*_m, \forall m$. We refer the reader to~\cite{bertsekas99,Komodakis2007} for details.

In order to specify our slave problems, we divide the set of voxels ${\cal V}$ into subsets ${\cal V}_m, m=1,\cdots,M$, such that
each pair of neighboring voxels $(i,j) \in {\cal N}$ appear together in exactly one subset ${\cal V}_m$.
Given such a division of voxels, our slave problems correspond to the following:
\begin{equation}
\min_{{\bf y}_m \in {\cal C}({\cal V}_m)} {\bf y}_m^\top L_m({\bf x};{\bf w}) {\bf y}_m + E_m^{\mathrm{prior}}({\bf y}_m,{\bf x};{\bf w}) + \rho^t_m {\bf y}_m,
\end{equation}
where $L_m({\bf x};{\bf w})$ is the Laplacian corresponding to the voxels ${\cal V}_m$. The prior energy functions $E_m^{\mathrm{prior}}$ 
modify the original prior $E^{\mathrm{prior}}$ by weighing each voxel $i \in {\cal V}_m$ by the reciprocal of the number of subsets
${\cal V}_n$ that contain $i$. In other words, the prior term for each voxel $i \in {\cal V}_m$
is multiplied by $1/|\{{\cal V}_n, i \in {\cal V}_n\}|$.

The slave problems defined above can be shown to provide a valid reparameterization of the original problem:
\begin{equation}
\min_{{\bf y} \in {\cal C}({\cal V})} {\bf y}^\top L({\bf x};{\bf w}) {\bf y} + E^{\mathrm{prior}}({\bf y},{\bf x};{\bf w}).
\label{eq:ACI}
\end{equation}
By using small subsets ${\cal V}_m$ we can optimize each slave problem in every iteration using a standard quadratic programming
solver. In our experiments, we used the Mosek solver. To the best of our knowledge, this is the first application of dual
decomposition to solve a probabilistic segmentation problem under linear constraints.

\bibliographystyle{splncs03}
\bibliography{learnRW,bib02.bib}
\end{document}